\documentclass[conference]{IEEEtran}
\IEEEoverridecommandlockouts
\usepackage{cite}
\usepackage{amsmath,amssymb,amsfonts}
\usepackage{tikz}
\usepackage{graphicx}
\usepackage{array}
\usepackage{colortbl}
\usepackage{algorithm}
\usepackage{textcomp}
\usepackage[utf8]{inputenc}
\usepackage{tcolorbox}
\usepackage{xcolor}
\usepackage{url}
\usepackage{booktabs} 
\def\BibTeX{{\rm B\kern-.05em{\sc i\kern-.025em b}\kern-.08em
    T\kern-.1667em\lower.7ex\hbox{E}\kern-.125emX}}
    \IEEEoverridecommandlockouts
    \usepackage{cite}
    \usepackage{amsmath,amssymb,amsfonts}
    \usepackage{algorithmic}
    \usepackage{graphicx}
    \usepackage{textcomp}
    \usepackage{xcolor}
    \def\BibTeX{{\rm B\kern-.05em{\sc i\kern-.025em b}\kern-.08em
    T\kern-.1667em\lower.7ex\hbox{E}\kern-.125emX}}
    \begin{document}

    \makeatletter
    \newcommand{\linebreakand}{%
      \end{@IEEEauthorhalign}
      \hfill\mbox{}\par
      \mbox{}\hfill\begin{@IEEEauthorhalign}
    }
    \makeatother

\title{VERITAS-NLI: Validation and Extraction of Reliable Information Through Automated Scraping and Natural Language Inference\\
}

 \author{
\IEEEauthorblockN{Arjun Shah$^{\dagger}$}
\IEEEauthorblockA{\textit{Computer Engineering} \\
\textit{D.J. Sanghvi College of Engineering}\\
arjun.a.shah244@gmail.com}
\and
\IEEEauthorblockN{Hetansh Shah$^{\dagger}$}
\IEEEauthorblockA{\textit{Computer Engineering} \\
\textit{D.J. Sanghvi College of Engineering}\\
shah.hetansh95@gmail.com}
\and
\IEEEauthorblockN{Vedica Bafna$^{\dagger}$}
\IEEEauthorblockA{\textit{Computer Engineering} \\
\textit{D.J. Sanghvi College of Engineering}\\
vedicabafna@gmail.com}
\linebreakand 
\linebreakand 
\IEEEauthorblockN{Charmi Khandor$^{\dagger}$}
\IEEEauthorblockA{\textit{Computer Engineering} \\
\textit{D.J. Sanghvi College of Engineering}\\
charmik42@gmail.com}
\and
\IEEEauthorblockN{Prof. Sindhu Nair}
\IEEEauthorblockA{\textit{Computer Engineering} \\
\textit{D.J. Sanghvi College of Engineering}\\
sindhu.nair@djsce.ac.in}
}

\maketitle
\begin{abstract}
In today’s day and age where information is rapidly spread through online platforms, the rise of fake news poses an alarming threat to the integrity of public discourse, societal trust, and reputed news sources. Classical machine learning and Transformer-based models have been extensively studied for the task of fake news detection, however they are hampered by their reliance on training data and are unable to generalize on unseen headlines. To address these challenges, we propose our novel solution, leveraging web-scraping techniques and Natural Language Inference (NLI) models to retrieve external knowledge necessary for verifying the accuracy of a headline. Our system is evaluated on a diverse self-curated evaluation dataset spanning over multiple news channels and broad domains. Our best performing pipeline achieves an accuracy of \textit{84.3\%} surpassing the best classical Machine Learning model by \textit{33.3\%} and Bidirectional Encoder Representations from Transformers (BERT) by \textit{31.0\%}. This highlights the efficacy of combining dynamic web-scraping with Natural Language Inference to find support for a claimed headline in the corresponding externally retrieved knowledge for the task of fake news detection.
\end{abstract}
{\def\thefootnote{}\footnotetext{$^{\dagger}$ These authors contributed equally to this work. }}
\section{Introduction}
As more and more people rely on digital platforms as their primary source of information, the threat of fake news has been exacerbated. This phenomena not only undermines the trust in reliable sources but also has the potential to skew public opinion~\cite{bib48}. The ease at which false narratives spread across social media and online platforms underscores the urgent need to address this issue~\cite{bib49}.

With the proliferation of fake news as evidenced by the spread of fabricated stories in the 2016 US elections~\cite{bib1}, individuals are often overwhelmed and unsure about the authenticity of the information they encounter.  With Google’s Adsense becoming the primary metric of success for a news site, a push has been made towards publishing ‘click-bait’ articles which catch a reader’s attention~\cite{bib2}. Zhou et al.~\cite{bib51} highlighted the explosive growth of fake news and it, consequentially leading to the erosion of democracy, justice, and public trust. Fake news is sometimes intentionally created and disseminated as part of misinformation campaigns aimed at manipulating public opinion or advancing specific agendas. Especially on social media, studies reveal that while Facebook has seen a decline in user interactions with misinformation since 2016, Twitter continues to see a rise~\cite{bib50}. The findings highlight that fake news and misinformation still prevail on social media platforms ~\cite{bib48}~\cite{bib49}. Hence ascertaining the veracity of  news articles will play a crucial role in preventing the spread of misinformation, thereby guaranteeing the populace's access to trustworthy and empirically-supported knowledge.

This paper introduces a novel solution, VERITAS-NLI, leverages state-of-the-art Natural Language Inference Models for verifying claims by comparing them against externally retrieved information from reputable sources in real-time via web scraping techniques. Furthermore, we employ small language models to generate questions based on the headline, enhancing the verification process through a question-answering approach. To assess the consistency between the scraped article and the input headlines, we employed NLI models of varying granularity. Additionally, we constructed an evaluation dataset by manually curating real headlines from reputable sources and synthetically generating fake headlines to simulate misinformation scenarios. These techniques enable our model to adapt to dynamic content, as it does not rely on static, outdated data. By continuously validating claims against real-time information, our approach ensures that the model remains relevant and effective in rapidly changing news environments.

Our model analyzes the content of news articles, posts, or other forms of media to identify potential indicators of fake news. This analysis involves examining the language, sources, factual claims, pronoun-swapping, sentence-negation, named-entity preservation and overall credibility of the information presented. The detector cross-references the claims made in the news content with the latest reliable and authoritative sources.

In the subsequent sections of this paper, we will delve into the underlying methodology of \textit{VERITAS-NLI} - Validation and Extraction of Reliable Information Through Automated Scraping and Natural Language Inference - elucidates the key features and algorithms employed, and presents empirical results demonstrating its efficacy and performance, which we have compared to the related work reviewed in section 2. We have conducted thorough comparisons by bench marking different approaches against our evaluation dataset, to analyze their performance in real-world situations. We have trained and compared multiple models from simple Machine-Learning approaches like Logistic Regression to more complex approaches utilising BERT and Natural Language Inference models, notably  FactCC~\cite{bib18} and SummaC~\cite{bib52}.
\vspace{1em}

\section{Literature Survey}
The initial attempts to counter false information relied on manually checking claims like the site factchecker.in~\cite{bib3}, known for it’s efforts to verify claims and statements made in the public domain. However, human fact-checking is resource-intensive and may not scale as easily to handle the sheer volume of information available online. Human fact-checkers may introduce subjectivity and potential bias in the interpretation of claims.

Recent advancement in machine learning have shown promise in improving the effectiveness of fake news detection. A significant number of studies have explored the use of classical machine learning algorithms for this purpose. Abdullah-All-Tanvir et al.~\cite{bib24} and Pandey et. al.~\cite{bib4} conducted comparative studies, both highlighting the higher performance of Support Vector Machine (SVM) and Naïve Bayes classifier in detecting fake news. Aphiwongsophon et al.~\cite{bib25}, further applied these methods to dataset collected from twitter API, reaching an accuracy of 96.08\%. However each of these studies relied on a niche focus on a specific topic, raising concerns about the broader applicability of these findings. Ahmed et al.~\cite{bib26} employed n-gram analysis along with machine learning techniques, again reporting SVM as the best performing model.

Other algorithms have also been utilized like Ni et al.~\cite{bib27} explored the use of Random Forest Classifier getting an accuracy of 68\% on PolitiFact and GossiCop datasets. Kesarwani et al.~\cite{bib28} used K-Nearest Neighbour (KNN) algorithm on a dataset collected from BuzzFeed News getting an accuracy of 79\%. Rajendran et al.~\cite{bib29} compared Decision Tree and Naive Bayes, finding that Decision Trees outperformed Naive Bayes and gave a mean accuracy of 99.70\% when tested on fake political news. Kotteti et al.~\cite{bib30} emphasized the importance of data preprocessing, particularly the handling of missing vales, while training classical machine learning models on the LIAR dataset. The effectiveness of ensemble methods have also been explored by researchers like Kaur et al.~\cite{bib32} who proposed a multi-level voting ensemble models that combined multiple classifiers based on their false prediction ratio. Bozuyla et al.~\cite{bib33} extended this concept by utilizing AdaBoost ensemble learning on top of Naive Bayes, improving the accuracy by 2\%. Many of the studies presented face limitations in achieving high accuracy or generalizability due to their reliance on event-specific datasets. While classical machine learning models can be used as valuable bench marking tools, they often fall short for the complex task of fake news detection highlighting the need for more advanced and robust models to improve accuracy and generalize findings across different contexts. 

Researchers at MIT~\cite{bib5} developed a deep learning network that detects patterns in the language of fake news. This model was trained on the data up to 2016 and is suggested to be used with other techniques like automated fact-checkers. Graph-based models like the one proposed by Zhang et al.~\cite{bib6} introduced a self-supervised contrastive learning and a new Bayesian graph Local extrema Convolution (BLC). This method aggregates node features in a graph while accounting for uncertainties in social media interactions. Despite achieving high accuracy on Twitter datasets, the model struggles with the feature learning challenges posed by the power-law distribution of node degrees in social networks. Another approach presented by Li et al.~\cite{bib7} introduces KGAT, a neural network utilizing graph attention networks, to perform fine-grained fact verification by analysing relationships between claim and evidence presented as a graph.

Transformer-based models like BERT have shown significant promise in this task of fake news detection. Alghamdi et al.~\cite{bib34} compared different machine learning and deep learning techniques and found that BERT-base outperformed other models using contextualized embeddings, demonstrating great efficacy across multiple datasets. This aligns with the findings of work done by Aggarwal et al.~\cite{bib35}, where BERT achieved an accuracy of 97.02\% on NewsFN dataset, outperforming LSTM and Gradient Boosted Tree models. Combating Covid-19 misinformation has  also been researched on, for example, Alghamdi et al.~\cite{bib36} in his other paper used BERT model on COVID-Twitter dataset getting an F1 score of 98\%, along with the work done by Ayoub et al.~\cite{bib37} who used DistilBERT and SHAP (Shapley Additive exPlanations). Arun et al.~\cite{bib38} reported a 96.5\% accuracy by analyzing headlines and the supporting text using BERT, and Kaliyar et al.~\cite{bib39} introduced FakeBERT, a model that employs bidirectional learning to enhance the contextual understanding of articles, contrasting with traditional unidirectional approaches of looking at a text sequence. Bataineh et al.~\cite{bib40} used Bidirectional Long Short-Term Memory (Bi-LSTM) which is further optimized by genetic algorithms.

In addition to transformers, hybrid models combining various deep learning techniques have been explored. Ajik et al.~\cite{bib47} employed a hybrid approach combining Convolutional Neural Networks (CNN) and Long Short-Term Memory (LSTM) networks, achieving an accuracy of 96\% on a dataset sourced from Kaggle. Dong et al.~\cite{bib42} emphasized the importance of contextual features and user engagement in the propagation of fake news. He combined supervised and unsupervised learning paths with CNNs, performing well even with limited labeled data. A similar approach was taken by Li et al.~\cite{bib43} where a self-serving mechanism was used to improve detection over traditional methods. Transformer based models like BERT have been largely used for fake news detection.

Techniques for natural language processing or NLP are essential for spotting linguistic clues that point to false information. Chesney et al.~\cite{bib8} brought attention to the difficulties in using current NLP techniques because incongruence must be identified in relation to the text it represents. Large language models like GPT-3 have also been used in studies like the one done by Farima et al.~\cite{bib9}, where they used GPT-3 for generating questions, responses and verifying facts. Despite promising results (77-85\% F1), has certain limitations like it depends on the initial set of questions generated by LLMs and operates on small-scale datasets, potentially limiting scalability and generalizability. We utilize open source Small Language Models (SLMs) and achieve comparable results on our evaluation dataset, presenting more quantitative findings. Another study by Eun et al.~\cite{bib10} employs LoRa for fine-tuning LLMs and evaluates claim accuracy across various claim styles by adjusting the temperature setting in the model. However, it majorly only focuses on combating fake news related to the coronavirus period, raising concerns about the generalizability of these claims.

The importance of social context in fake news detection has been highlighted in various studies. Research done by Shu et al.~\cite{bib44}talked about the challenges in understanding the underlying characteristics of fake news, suggesting more research on utilizing social context. Kang et al.~\cite{bib45} further emphasized the importance of incorporating user engagement metrics, showing that user information can enhance detection even when data contains hate speech.

There have been suggestions to use hybrid approaches that combines both linguistic and network-based methods like the work done by Conroy~\cite{bib11} who presented a comprehensive survey of current technologies and methodologies for detecting fake news. They emphasised that tools should be designed to augment human judgement, not replace it.

Several datasets and evaluation metrics have been developed to facilitate research in fake news detection. The FEVER dataset, introduced by Thorne et al.~\cite{bib12}, serves as a benchmark for fact extraction and verification tasks. Hassan et al.~\cite{bib13} made a data collection site to collect ground-truth labels of the sentences and deployed a automated fact-checking system on the 2016 presidential primary debate and compared its performance to professional news and organisations. Reddy et al.~\cite{bib14} leveraged ClaimBuster model for claim detection and employed a Bart-large model trained on a MNLI (MultiNLI) corpus as zero-shot topic-filtering system to filter out claims related to the topics under consideration.

In summary, the research on fake news detection has shown significant progess in developing various methods and models. However, several research gaps have been observed as noted by Almahadi et al.~\cite{bib46} in his comprehensive review of different approaches of fake news detection (FND) . One notable limitation is the lack of real-world implementation of these techniques. While theoretical models have been explored extensively, there is a need to address practical issues related to scalability, robustness, and the ethical implications of implementing FND in the real-world scenario. Moreover, exploration of novel ML algorithms specifically for FND that address the limitations of classical ML algorithms, such as their reliance on static training data and inability to keep up with the current news in a dynamic environment. Our work aims to tackle both of these areas, moving closer to a robust and practical solution for combating fake news.

\vspace{1em}

\section{Methodology}
\subsection{Training Dataset}\label{AA}
In our study, we utilize the LIAR dataset , a publicly available resource for fake news detection compiled from POLTIFACT.COM, a Pulitzer Prize-winning website. The credibility of this dataset can be attributed to the multiple human evaluated statements along with authentication of the POLITIFACT editor~\cite{bib30}~\cite{bib34}~\cite{bib42}.

The dataset samples news claims from various multimedia sources ranging from news releases, TV/radio interviews, campaign speeches, advertisements and social media platforms such as Twitter, Facebook etc. This dataset primarily focuses on societal and government related news topics ranging from economy, taxes, campaign-biography, elections, education, jobs are some the most represented subjects. The claims based on these subjects are the most prone to misinterpretation, and spreading of misinformation by tweaking minuscule sections of the claim to drastically change the implication of the claim. This serves as strong foundation for the culmination of a robust fact and integrity analyzer developed on the foundation of classical NLP Techniques and cross-verification using web-scraping methodologies~\cite{bib57}~\cite{bib58}.

The LIAR dataset consists of 12.8k statements labeled for truthfulness, collected over a ten-year period. Each statement is accompanied by metadata such as the speaker, context, and a detailed judgment report. The LIAR dataset is structured to include several fields that capture various dimensions of each statement recorded. Here’s an elaboration on each type of field found in the LIAR dataset:
\\
\begin{itemize}
    \item\textit{Label Description:} The truthfulness rating assigned to the statement. LIAR dataset uses a multilabel system, which can include labels such as "True", "Mostly True", "Half True", "Barely True", "False", and "Pants on Fire".
    \item\textit{Label Assignment}: The rationale or explanation is provided by fact-checkers on why a particular label was assigned to the statement. This field often includes evidence or reasoning based on research, reports, and data analysis. The dataset consists of a count for each of the labels, for a particular statement. The aggregate of these labels are compared and the claim is thus assigned a primary label for representation.
\end{itemize}

We convert these 6 classes into two distinct categories (True/False) to abate the ambiguity posed by these 6 varying classes. By reducing the classes to binary outcomes, we emphasize the fundamental distinction between accurate and inaccurate information. Additionally, the binary categorization mitigates subjective disagreements that may arise from the nuanced differences between the original labels, adhering to consistent labeling standards.
\\
The label mapping we used for the conversion is:
\begin{verbatim}
Mapping = {
    'true': True,
    'mostly-true': True,
    'half-true': True,
    'barely-true': False,
    'false': False,
    'pants-fire': False
    }
\end{verbatim}
\vspace{0.15cm}
We made use of the entire transformed LIAR dataset for the training of classical machine learning models and transformer based models for binary classification.
\begin{figure}[ht]
    \centering
    \includegraphics[width=0.9\linewidth]{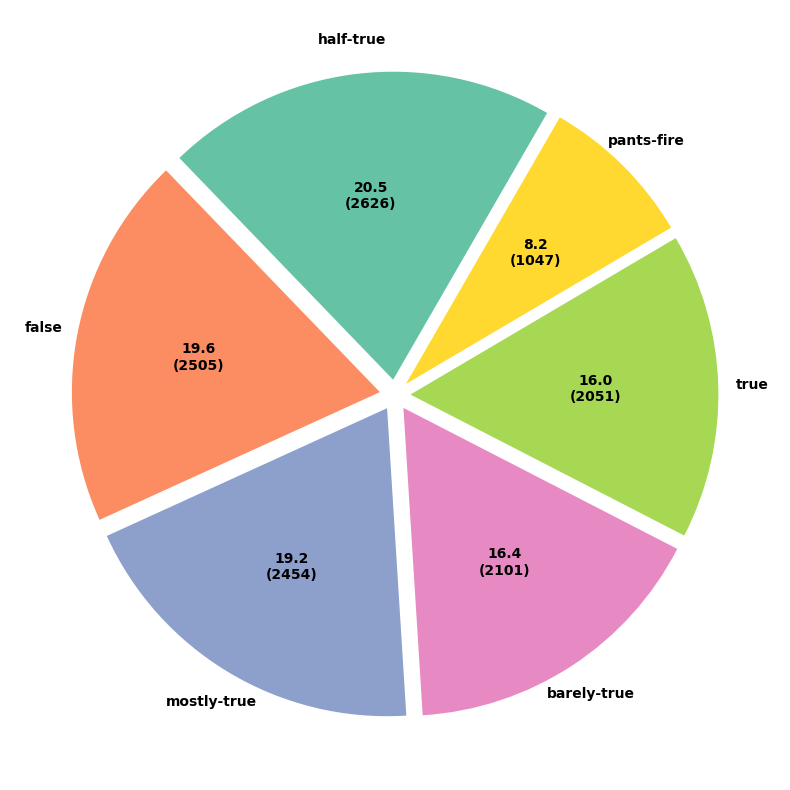}
    \caption{The figure presents the label distribution of the LIAR dataset. Labels categorize claims as "True", "Mostly True", "Half True", "Barely True", "False", and "Pants on Fire". The label denotes the credibility of each headline in the dataset.}
\end{figure}
\subsection{Evaluation Dataset}\label{BB}
To evaluate the performance of the models trained, we propose a new manually curated dataset consisting of the latest headlines of 2024. This dataset with more recent headlines help us benchmark the efficacy of the various models trained in identifying unreliable headlines in a dynamic and rapidly-evolving landscape.

The dataset consists of headlines from a wide range of domains, including science, sports, pop culture, politics, and business. 
It contains 346 credible headlines collected from various trusted sources such as CNN, USNews, The Guardian, BBC, The New York Times, Reuters, Times of India, CNBC, and others.

Using each reliable headline, we generate a corresponding unreliable headline using a Small Language Model(SLM), specifically \texttt{microsoft/Phi-3-mini-4k-instruct}. We devised a zero-shot prompt without any examples, resulting in a total of 692 headlines in our evaluation dataset. 

The small language models were prompted to modify the reliable headlines through a combination of techniques such as sentence negation, number swaps, replacing named entities, and by injecting noise into the headline. The aim was to create convincing fake news headlines that can serve as a challenging evaluation metric for the models trained.
\\\\
\textit{Unreliable Headline Generation Prompt}:
\texttt{"Given the true headline ,you have to make changes to it using a combination of Sentence negation, Number swaps, Replacing Named Entities and Noise Injection to generate a fake news headline. The output should be only the generated fake news headline to evaluate my fake news model."}

\begin{figure}[ht] 
    \centering
    \includegraphics[width=0.9\linewidth]{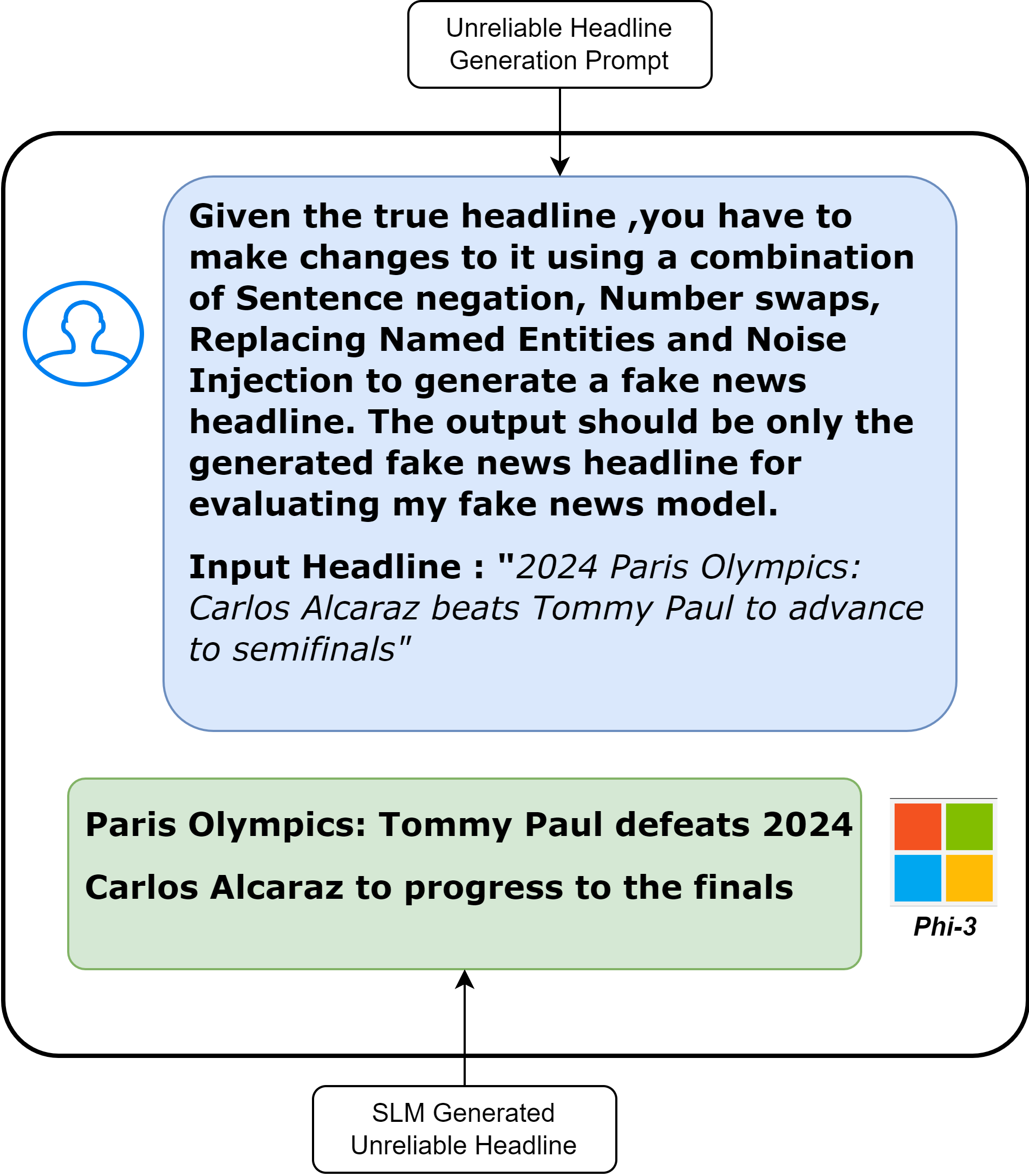} 
    \caption{Example of an Unreliable Headline generated using Microsoft's Phi-3, when prompted with our zero-shot prompting approach.}
    \label{fig:PhiQuestion}
\end{figure}
\subsection{Classical Model Performance Benchmarks}

\subsubsection{\textbf{Classical Machine Learning}}\label{CC}
Classical Machine Learning algorithms have been used extensively for the task of analyzing the veracity of news articles~\cite{bib4} , capitalizing on the wealth of data provided by datasets like LIAR to train models.
\\
Prior to feeding the data from the LIAR dataset for the training of these models, a series of pre-processing steps have to be taken.
\begin{itemize}
    \item \textit{Stopword Removal}: In this step extremely common words (known as stopwords),that carry little meaning for the task of prediction, are removed from the text to reduce the noise in the dataset.
    \item \textit{Token Filtering}: we perform token filtering prior to lemmatization in order to exclude non-alphabetic tokens. This step ensures that the non-alphanumeric characters do not impede model training. 
    \item \textit{Lemmatization}: Words in the news articles and headlines are reduced to their root form to normalize the text.This is done using NLTK's WordNetLemmatizer.
    \item \textit{TF-IDF Vectorizer}: TF-IDF stands for Term Frequency Inverse Document Frequency.It is a mathematical method that is used to represent how important a word is within a document relative to a collection or corpus. This importance of a word is directly proportional to how often it appears in a document but is offset by the frequency of the word in the complete text corpus. 
    \\
    \textit{Term Frequency (TF)}
    The term frequency \( tf(t, d) \) is defined as:
    \[
    tf(t, d) = \frac{f_{t, d}}{\sum_{t' \in d} f_{t', d}}
    \]
    where \( f_{t, d} \) is the frequency of term \( t \) in document \( d \), and the denominator is the sum of all term frequencies in that document.  
    \textit{Inverse Document Frequency (IDF)}
    The inverse document frequency \( idf(t, D) \) is defined as:
    \[
    idf(t, D) = \log \left(\frac{N}{|\{d \in D : t \in d\}|}\right)
    \]
    where \( N \) is the total number of documents in the corpus \( D \), and \( |\{d \in D : t \in d\}| \) is the number of documents where the term \( t \) appears.
\end{itemize}
\vspace{1em}
Using the aforementioned preprocessing techniques, we trained the following models:
\begin{itemize}
    \item \textit{Linear SVC}: Linear Support Vector classifier is a base classical model which attempts to find a hyperplane to maximize the distance between classified samples. It requires and functions around a kernel which provides adaptations for dealing with non-linear data-points.~\cite{bib4}~\cite{bib32}.
    \item \textit{Multinomial Naive-Bayes}: This classical model belongs to the family of probabilistic classifiers derived from the eminent Bayes Theorem. Multinomial Naive Bayes is a variant of Naive Bayes used for document classification which makes it a desirable choice to use for our particular problem. Each sample represents a feature vector in which certain events have been counted in the multinomial model~\cite{bib4}~\cite{bib29}~\cite{bib33}. 
    \item \textit{Random Forest Classifier}: This model represents a meta-estimator creating a number of decision tree classifiers considering sub-samples of words in the pre-processed dataset to improve predictive accuracy~\cite{bib27}.
    \item \textit{Logistic Regression}: It is used to model the relationship between one dependent and one or more independent variables. This model predicts the likelihood of something falling into a specific group. It uses the logistic function to map the linear combination of independent variables to a probability between 0 and 1 for each class.~\cite{bib24}~\cite{bib26}~\cite{bib32}~\cite{bib44}.
\end{itemize}

\subsubsection{\textbf{Bidirectional Encoder Representations from Transformers}}
The Liar Dataset consists of very comprehensive claims, for which the labels naturally rely heavily on nuanced sentence interpretation and context. Owing to the performance demonstrated by transformers in capturing textual information we naturally diverted our attention to transformer based models. Fine-tuning pre-trained models such as BERT-base-uncased has allowed us to explore an architecture capable of identifying discrepancies in general news patterns due to it's deeper understanding of context through vector representations of characters in the English language stemming from the fact that it has been trained on a large corpus of over three billion words~\cite{bib16}.

BERT is based on the Transformer~\cite{bib19} architecture, which captures the context of words in a sentence by considering words from both left and right contexts. This is essential for language tasks like fake news detection, where understanding the context in which words appear can determine the authenticity of the information~\cite{bib34, bib35, bib36, bib37, bib38,bib39}.
BERT is pre-trained on "masked language modeling" and "next sentence prediction" and for our specific use-case, "sequence classification". This pre-training helps the model learn a wide range of language patterns and knowledge, which might not be present in the specific training dataset, however, fine-tuning it on our dataset using transfer-learning can provide an advantage while solving a specific problem such as fact-checking. For the fine-tuning process, task-specific layers are added to the BERT architecture for the downstream sequence-classification task, and only these newly added layers are trained to minimize computational costs.
\begin{table*}[ht]
\centering
\caption{Classical Machine Learning Model Performance Metrics for LIAR and Evaluation Test Sets\\}
\begin{tabular}{|>{\centering\arraybackslash}m{3cm}|>{\centering\arraybackslash}m{2.5cm}|>{\centering\arraybackslash}m{2.5cm}|>{\centering\arraybackslash}m{2.5cm}|>{\centering\arraybackslash}m{2.5cm}|}
\hline
\textbf{\textit{Model}} & \textbf{\textit{Accuracy}} & \textbf{\textit{Precision}} & 
\textbf{\textit{Recall}} & \textbf{\textit{F1-Score}} \\ \hline
\multicolumn{5}{|c|}{\textbf{Models trained and tested on LIAR dataset}} \\ \hline
LinearSVC & 59.34\% & 0.64 & 0.66 & 0.65 \\ \hline
Logistic Regression & 60.52\% & 0.63 & 0.72 & 0.68 \\ \hline
MultinomialNB & 60.54\% & 0.61 & 0.83 & 0.70 \\ \hline
Random Forest Classifier & 59.76\% & 0.63 & 0.71 & 0.67 \\ \hline
\multicolumn{5}{|c|}{\textbf{Models trained on LIAR and tested using the Evaluation dataset}} \\ \hline
LinearSVC & 49.13\% & 0.49 & 0.49 & 0.49 \\ \hline
Logistic Regression & 50.00\% & 0.50 & 0.50 & 0.54 \\ \hline
MultinomialNB & 51.02\% & 0.51 & 0.64 & 0.57 \\ \hline
Random Forest Classifier & 49.27\% & 0.49 & 0.62 & 0.55 \\ \hline
\multicolumn{5}{|c|}{\textbf{Model trained on LIAR and tested using the Evaluation dataset}} \\ \hline
BERT-Base-Uncased & 53.34\% & 0.52 & 0.71 & 0.60 \\ \hline
\end{tabular}
\end{table*}

BERT's bidirectional context understanding helps it effectively handle polysemous words (words with multiple meanings) better than models that consider only one-directional context or no contextual information at all. This is important because within polysemous words and the words that come before and after them, the severity of the statement is identified on varying scales. For example, the negative connotation of a word in a particular context is enhanced based on the synonym of the word used. Unlike classical models that often require significant customization and feature engineering, BERT can be fine-tuned with a small number of task-specific examples to achieve high performance, leveraging the extensive pre-training.
The procedure involving the fine-tuning of a BERT model on the LIAR dataset begins with an intricate tokenization process. For this process, we use the WordPiece algorithm~\cite{bib20} :
\begin{enumerate}
    \item \textbf{Text Normalization:} Convert all text to lowercase (for BERT-base-uncased) and perform unicode normalization.
    \[
    \text{Text} \rightarrow \text{Normalized Text}
    \]
    \item \textbf{Punctuation Splitting:} Split the text at punctuation, adding spaces around each punctuation mark.
    \[
    \text{Normalized Text} \rightarrow \text{Split Text}
    \]
    \item \textbf{WordPiece Tokenization:} Break words into subword units based on a trained vocabulary. Unknown words are split into smaller subwords that exist in the vocabulary.
    \[
    \text{Word} \rightarrow \text{Subword}_1 + \text{Subword}_2 + \ldots + \text{Subword}_n
    \]
    \item \textbf{Adding Special Tokens:} Add special tokens such as \texttt{[CLS]}, \texttt{[SEP]} for specific tasks (e.g., classification, question answering).
    \[
    \text{Final Tokens} = \text{[CLS]} + \text{Tokenized Text} + \text{[SEP]}
    \]
\end{enumerate}

This process enables BERT to handle a wide range of language inputs and effectively manage both common and rare words.
This is done along with padding and truncation to ensure equal length input vectors for uniformity. We employ transfer learning techniques using the BERT-base-uncased model, sourced from the Hugging Face transformers library. The methodology focuses extensively on the fine-tuning process:
\begin{figure}
    \centering
    \includegraphics[width=1\linewidth]{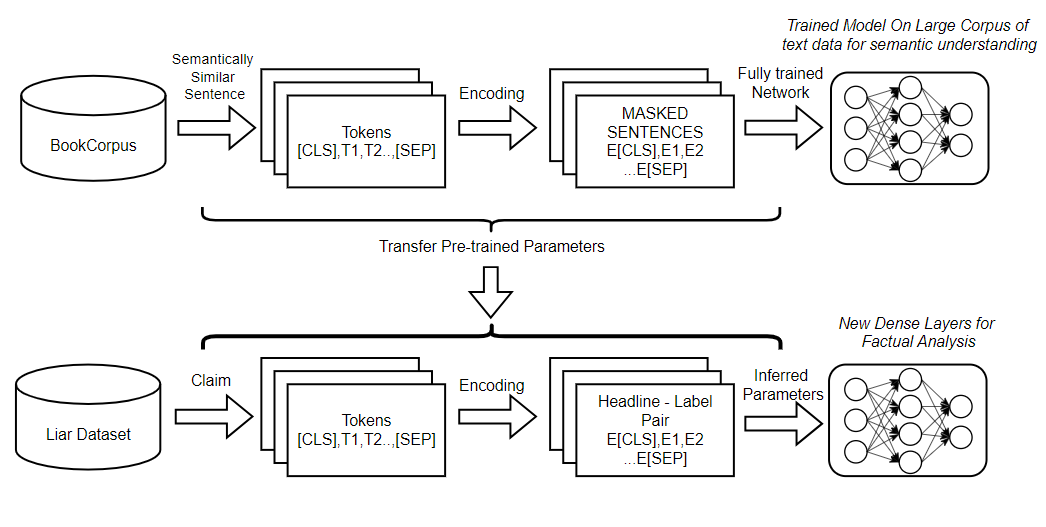}
    \caption{Fine-Tuning of BERT using transfer learning utilizing the LIAR dataset to ascertain the veracity of claimed headlines.}
    \label{fig:TransferLearning}
\end{figure}

\begin{itemize}
        \item \textbf{Initialization:} Fine-tuning begins with the pre-trained models which have been initially trained on a large corpus, such as the BooksCorpus and English Wikipedia. This pre-training has equipped the models with a robust understanding of general language contexts. 
        \item \textbf{Task-Specific Adjustments:} The models are then further fine-tuned on a labeled dataset specific to our use case. This process involves tweaking the last few layers of the model so that the model can align its outputs with the task at hand.
        For our requirement of binary-classification of fake news using the LIAR dataset, it entails training the model on the labeled dataset, consisting of news articles, headlines and labels.
        This adjustment allows the model to conform its outputs to align with the specific requirements of binary classification.
        \\
        \item \textbf{Hyperparameter Optimization:} 
        \begin{itemize}
        \item \(num\_train\_epochs = 1\): Given the high representational power of BERT and its pre-trained nature, a single epoch is often sufficient to fine-tune the model. Experimental results found that using a greater number of epochs lead to model-overfitting.
        \item  \begin{flushleft}
        \(per\_device\_train\_batch\_size \) = 16 and \(per\_device\_eval\_batch\_size\) = 32: Smaller batch sizes for training help in maintaining a balance between memory usage and effective learning.
        \end{flushleft}
        \item \(warmup\_steps = 100\): Implementing a warmup phase at the beginning of training where learning rates are gradually increased helps in stabilizing the learning process, preventing the model from converging too quickly to a sub-optimal solution.
        \item  \(learning\_rate = 5e-5\): This learning rate is found to be most suitable for many models for the Adam Optimizer which was used in this particular model training.
        \item \(weight\_decay = 0.001\): This helps in regularizing the model and preventing over-fitting, which is crucial for a model as large as BERT.
        \item \(fp16\ =\) True: This enables the model to train faster and consume less memory while maintaining the training precision, making it feasible to train larger models or use larger batch sizes.
        \end{itemize}
    \end{itemize}
\section{Proposed Solution }\label{DD}
Our proposed solution, VERITAS-NLI explores various novel pipelines that make use of web-scraping, Small Language Models (SLM) and Natural Language Inference (NLI) models at its core. This solution mimics the natural cognitive processes humans use to evaluate whether a particular claim is reliable or unreliable. The pipelines, namely: \texttt{Question-Answer Pipeline, Article Pipeline and Small Language Model Pipeline} are a testament to this underlying intuition and demonstrate it's effectiveness in the subsequent sections.
\subsection{Web Scraping Techniques}
We make use of web-scraping techniques to retrieve external knowledge based on the raw input headline, the questions generated by small language models to verify the headline, the 'People Also Ask' section of a Google search and  Google's 'Quick Answer' to retrieve relevant information in real-time. This real-time knowledge is critical for our task of detecting fake news in a dynamic and fast-changing environment~\cite{bib58, bib59}.\\
\textit{Selenium Based Scraping}:
Our pipeline leverages a 'chromedriver' and adaptive web-scraping of the document-object-model(DOM) by traversing the XML Tree to acquire external information to aid in verifying the input~\cite{bib57}. There are multiple different approaches we take to scrape external information relevant to the headline, they are outlined below :
\begin{itemize}
    \item \textbf{Fact-Based Headline Verification: }For user headlines that represent facts we can verify them by scraping the text that is provided by the Google Search 'Quick Answer'. These webpages do not change their content and layout frequently and are known as static pages which allows us to simply retrieve the Quick Answer by scraping that HTML tag represented in the XML tree.
    \item \textbf{Top-K Articles Retrieval: }A naive google search is performed initially to retrieve the top-k links corresponding to the headline. This is followed by the scraping of relevant 'titles' and useful 'contents' from these links (generally represented by \texttt{<h>} and \texttt{<p>} html tags) to cross-validate the claimed headline.
    \item \textbf{People Also Asked: }We incorporate Google Search's 'People Also Asked' (PAA) questions relevant to the news-headline entered by the user. The People Also Asked question provides different interpretations for a particular headline most frequently inferred by people. This also provides an all-rounded context for the evaluation of the claimed headline. Additionally, more popular claims will consist of a greater number of questions which will inherently provide a more comprehensive context that assists the model to identify confidence in a given claim as well as sometimes point out the ambiguity of claims that are opinionated, unclear, or shrouded in noise.
    \item \begin{flushleft}
    \textbf{SLM Generated Questions:}
    Furthermore, we make use of Small Language Models (\texttt{Mistral-7B-Instruct\\-v0.3}~\cite{bib22} or Microsoft's \texttt{Phi-3-mini-4k-\\instruct}~\cite{bib23}) to generate a question that helps us retrieve relevant information needed to validate the reliability of a given headline.
    \end{flushleft}
\end{itemize}

 We utilize trusted sources for information retrieval and adhere to the 'robots.txt' laid down by all of the sources such as CNN, USNews, The Guardian, BBC, The New York Times, Reuters, Times of India, CNBC, etc which scrutinize the published articles on their respective platforms and are generally regarded as trustworthy news outlets.
 
A 'robots.txt' file is a directive used by website administrators to communicate with search engine crawlers and web scrapers, specifying which pages or sections of a site they are permitted to access and index. Its primary purpose is to manage web traffic, ensuring that crawlers do not overwhelm the server with excessive requests. By controlling the areas of the site accessible to bots, the 'robots.txt' file helps optimize server performance, protect sensitive information, and prioritize which content should be visible in search engine results.
\subsection{Question-Answer Pipeline}
In this pipeline we experiment with an amalgamation of methods using the Google Search 'Quick Answer' which is highly dependent on whether the given input headline is a hard fact or not. So to make this a more thorough pipeline we've incorporated the  ’People Also Asked’ (PAA) questions for the input headline, as a fallback to the 'Quick Answer' scraping. However, we found that the 'People Also Asked Section' of a particular Google page is heavily dependant on user interaction for a particular headline if the input is an uncommon occurrence it may not be represented by the 'People Also Ask' section. In this case we have a terminating fallback which resorts to the naive scraping of the top-k articles that are found. 

The third pipeline in Figure 5, highlights the flow of the web-scraping and claim verification process. The scraped content is then passed to the headline-verifying NLI model as the source-text (premise) and the input headline is the claim (hypothesis). The model then generates a prediction along with a confidence score which represents how well the claims made in the headline are supported by the externally scraped information.

\subsection{Small Language Model Pipeline} 
Within this pipeline, we explore the use of Small Language Models to generate questions that can be used scrape more relevant content to verify the entered claim. We follow a scraping approach identical to the one used in the Question-Answer Pipeline with an added zero-shot\cite{bib31} SLM question generation step \cite{bib9}.

This pipeline involves the transformation of the user entered input headline into a context-aware question using an Open Source Small Language Model, like \texttt{Mistral-7B-Instruct-v0.3}\cite{bib22} and Microsoft's \texttt{Phi-3-mini-4k-instruct}\cite{bib23} which have 7.25 and 3.82 billion parameters respectively, these 
 questions help us retrieve more direct answers from the web to fact-check claims against external sources. Leveraging the capabilities of a Small Language Model for question generation should likely assist us in retrieving more pertinent information helping verify the headline.
\\
\begin{figure}[ht]
    \centering
    \includegraphics[width=0.9\linewidth]{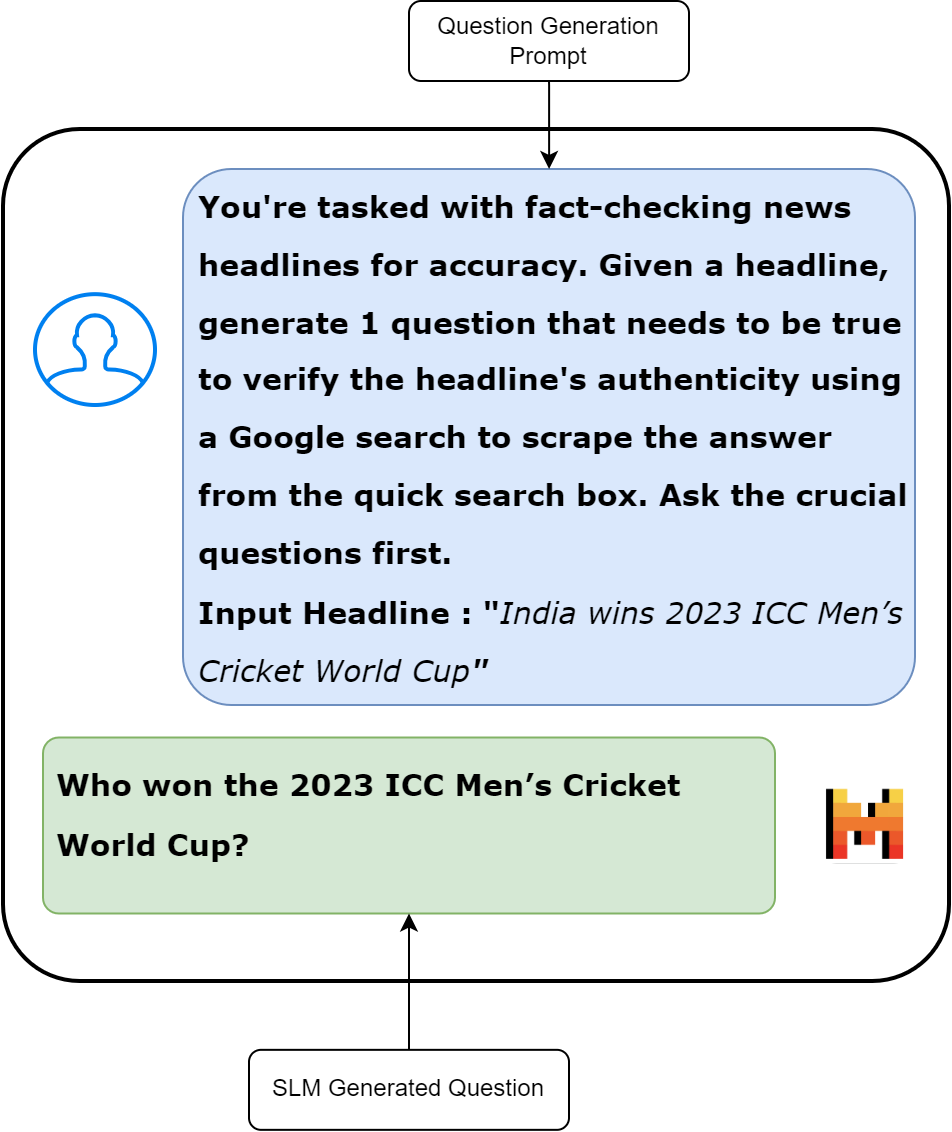} 
    \caption{Example of a question generated by Mistral 7B to aid in the knowledge retrieval, a crucial step of the Small Language Model Pipeline.}
\end{figure}

\begin{figure*}
    \centering
    \begin{tikzpicture}
        \node[draw=black, line width=0.1mm, rectangle, inner sep=10pt] (image) at (0,0) {
            \includegraphics[width=1\linewidth]{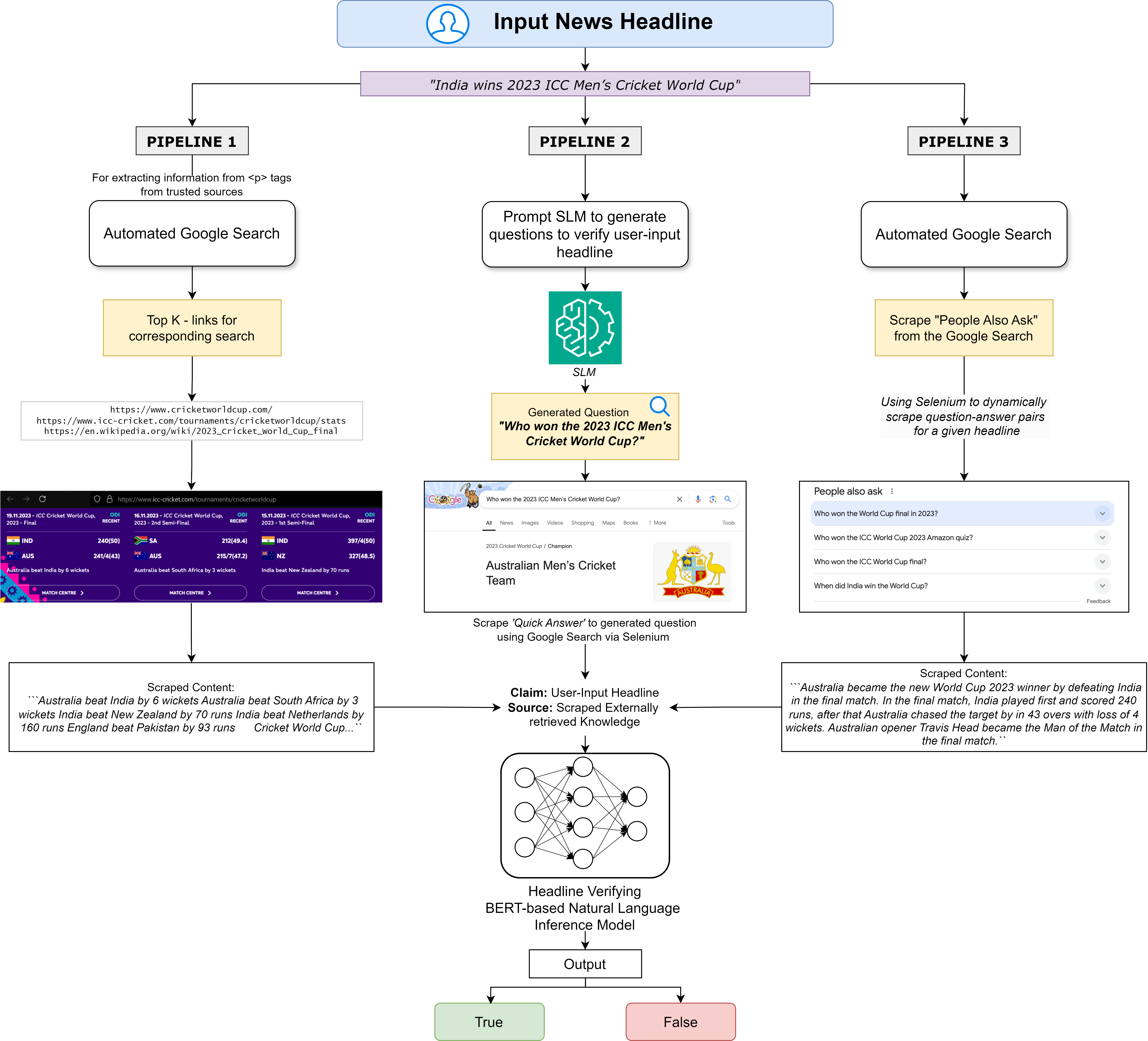}
        };
    \end{tikzpicture}
    \caption{Architecture Diagram of our proposed solution VERITAS-NLI, detailing the workflow of our 3 proposed pipelines.}
\end{figure*}
Figure 4 illustrates the question-generation process using the input headline \textit{'India wins 2023 ICC Men's Cricket World Cup'} along with a zero-shot prompt and the question generated by the SLM is \textit{'Who won the 2023 ICC Men's Cricket World Cup?'}, a question like this significantly increases the probability of obtaining a direct response on the web.
\\
\textit{Example Question Generation Prompt :}
\texttt{You're tasked with fact-checking news headlines for accuracy. Given a headline, generate 1 question that needs to be true to verify the headlines authenticity using a Google search to scrape the answer from the quick search box. Ask the crucial questions first.}
\\
Pipeline 2 in Figure 5 illustrates the question generation process for an input headline, by prompting \texttt{Mistral-7B-Instruct-v0.3}. This generated question is then used for the subsequent knowledge retrieval from the 'Quick-Answer' section of Google Search. The generated question simplifies the scraping workflow and helps retrieve a direct answer to verify the veracity of the claim.

\subsection{Article Pipeline}
The article pipeline leverages a web-scraping methodology~\cite{bib57} to gather data from various online sources systematically. This approach allows for the extraction of relevant content, such as content related to a particular input headline, based on the Top - K links to articles retrieved as illustrated in Pipeline 1 of Figure 5. The web scraper acquires both the headings, and the content from these links and the model is supplied with this retrieved information for the verification process. By structuring the pipeline in this way, it ensures that the process is both efficient and reliable, the inclusion of an article's content helps provide a rich-contextual understanding of the topic at hand to help verify the claim.

Interestingly, the research has shown that this naive approach leads to more accurate outcomes. In this context, a simple yet well-organized strategy for scraping and processing data performs better than more intricate alternatives. To contrast this pipeline intimately with the previous three where ‘Quick Answer’ and ‘People Also Asked’ summaries are provided to the model may be small or incomplete for certain user queries misguiding the model. The best output is closely correlated with the scraping of the right amount of information quantity to quality (coherence) of the scraped content. This is further highlighted from the statistical results obtained and delineated in subsequent results section.

The scraped articles pertinent to the claim, are then processed by Natural Language Inference (NLI) models, specifically FactCC~\cite{bib18} and SummaC~\cite{bib52}, that make use of the externally retrieved knowledge to analyze and detect potential inconsistencies within the claimed headline.
\subsection{NLI-based Models for Headline Inconsistency Detection}
The next step in our proposed solution is to detect inconsistencies in the headline (hypothesis) by comparing it with the source (premise) text fetched from the web. This step is analogous to the task of Natural Language Inference (NLI).
\\
These NLI models are designed to determine the relationship between two pieces of text, a 'premise' and a 'hypothesis', and are classified into 3 categories:
\begin{itemize}
    \item \textbf{Entailment:} The hypothesis is logically supported by the premise.
    \item \textbf{Contradiction:} The hypothesis is logically inconsistent with the premise.
    \item \textbf{Neutral:} The hypothesis neither entails nor contradicts the premise.
\end{itemize}
The performance of these models has seen substantial improvements in recent years, supported by large training datasets like MNLI~\cite{bib53}, SNLI~\cite{bib54} and VitaminC~\cite{bib55}. Modern attention-based architectures exhibit proficiency that parallels human performance for this task.~\cite{bib52}
\subsubsection{FactCC}
One approach of our proposed solution, VERITAS-NLI leverages FactCC~\cite{bib18} , a model developed to ensure that summaries generated by abstractive models remain factually consistent with their source documents~\cite{bib18}. This section delves into the architecture of FactCC and its contributions to the field of text summarization.
FactCC is built upon the BERT architecture, leveraging its pre-trained powerful contextual embedding capabilities. The model operates under a weakly-supervised learning framework, which is designed to handle the challenges of factual consistency in text summarization. It was trained on synthetic data which enables it to be a perfect candidate for downstream learning tasks.
\\
FactCC employs a \textit{multi-task}~\cite{bib21} learning approach, where the model is simultaneously trained on several related tasks:
\begin{itemize}
    \item \textbf{Consistency Identification:} Determining if a transformed sentence from the source document remains factually consistent.
    \item \textbf{Support Extraction:} Identifying and extracting the specific spans in the source document that support the factual consistency of the summary.
    \item \textbf{Error Localization:} Pinpointing and highlighting the spans within the summary that are factually inconsistent.
\end{itemize}
\begin{table*}[ht]
\centering
\caption{Proposed Solution Output for a Headline}
\begin{tabular}{|>{\raggedright\arraybackslash}p{2cm}|>{\raggedright\arraybackslash}p{4cm}|>{\raggedright\arraybackslash}p{7cm}|>{\raggedright\arraybackslash}p{2cm}|}
\hline
\rowcolor[gray]{0.9}
\textbf{Type of Content} & \textbf{Headline and Scraped Questions} & \textbf{Retrieved Article/Direct Answer} & \textbf{NLI-Model Output Score} \\ \hline
\multicolumn{4}{|c|}{\textbf{Input Headline : Max Verstappen wins 2023 F1 world title}} \\ \hline
Phi-3 Generated Question & Who won the 2023 Formula 1 World Championship? & Max Verstappen  & 0.9992 (CORRECT) \\
\hline

People Also Asked & Has Verstappen won the World Championship in 2023? & Has Verstappen won the World Championship in 2023? F1's most races left in a season before championship win.It became clear very early on that Verstappen would become a triple world champion in 2023. His season has been that dominant that the Red Bull driver sealed the championship in October after the Qatar GP sprint race.& 0.9968 (CORRECT)  \\
\hline

Scraped Article & Max Verstappen crowned Formula One world champion & Max Verstappen crowned Formula One world champion Max Verstappen was crowned Formula One world champion for the third time, securing the title in the sprint race at the Qatar Grand Prix on Saturday.There are few outcomes in sport that seem inevitable. Human and, in F1, engineering error can strike at any moment, with competitors waiting to pounce. But occasionally, someone comes along and dominates in a way that any anticipation, any lingering tension regarding the outcome dissipates Verstappen’s coronation  & 0.8776 (CORRECT)\\
\hline
\end{tabular}
\end{table*}
\subsubsection{SummaC (Summary Consistency)}
It obtains state-of-the-art results for the task of inconsistency detection in summarization of documents. The fundamental aspect that sets SummaC~\cite{bib52} apart is its segmentation of longer documents into sentence-level units, thus changing the granularity at which NLI models are applied.
\\\\
For each input \textit{(document,summary)}
\begin{itemize}
    \item The summary\textit{(S)} is split into \textit{N} sentence segments and similarly the Document\textit{(D)} is split into \textit{M} segments.
    \item For every such summary and document-sentence pair, the NLI model calculates the likelihood of three possible relationships (Entailment, Contradiction and Neutral), this results in a pair matrix.
   \item \textbf{SummaC-ZS} : The pair matrix is then reduced to a one-dimensional vector by taking the maximum value of each column. The mean of this vector is then taken to produce the final output score.
    \item \textbf{SummaC-Conv} : The intuition behind SummaC-Conv is to be able to capture NLI scores generated over the pair matrix instead of the extremas as was in the case of SummaC-ZS. A histogram of the  distribution is constructed on which binning is performed. This binning matrix is then passed through a 1D-Convolution. The convolution layer scans the summary histograms one at a time, and compiles each into a scalar value for each summary. 
    \item The task at hand requires our models to output binary predictions, thus for our SummaC models an optimal decision threshold is required to generate binary classification outputs. The threshold is determined through a calibration process utilizing 20\% of the evaluation dataset, with thresholds in the range of [-1,1]. The computation of this threshold is in accordance with the original study that proposed the SummaC architecture~\cite{bib52}.
\end{itemize}
\section{ Results}
In our comprehensive study on the efficacy of machine learning and deep learning models for fake news detection, our study finds significant performance differences between classical Machine Learning models, BERT models and our proposed Transformer-based pipelines that allow it to access external knowledge in order to classify the news headline at hand.

Classical Machine Learning models rely heavily on static training data, this dependency severely hampers their ability to classify news headlines in a dynamic environment. These straightforward approaches cannot adapt effectively to ever-changing relationships and contexts inherent in real-world scenarios. 

From the results in Table I, the accuracy's of classical machine learning models for the given binary-class classification task of news headlines on the training dataset produce an average F1-score of ~0.675 and an accuracy of ~60.04\% when tested on the LIAR dataset using a 70:30 train-test split. 

However, when these models were trained entirely on the LIAR dataset and then tested on our new proposed evaluation dataset, they produce an average F1-score of ~0.537 and accuracy of 49.86\%. This sharp decline highlights the models' reliance on static training data, which limits their ability to accurately classify unseen news articles in a dynamic and ever-changing environment. The drop-off in performance is exacerbated by the fact that the models tested on our evaluation dataset were trained on a larger corpus of news headlines, due to the lack of a train-test split.

Thus, the inability of these models to adequately generalize from their training data to accurately classify current headlines, renders them unsuitable for the task at hand. Nevertheless, these classical models serve as an important  baseline for subsequent, more advanced models.

To address the above limitations, our focus turns to a fine-tuned BERT, a transformer-based model, for binary classification of news articles. The process of fine-tuning these transformer models for our specific task empowered this study to explore an architecture with the ability to the understand the true-context within nuanced headlines. 

Although these models showed initial promise they also suffered the same pitfalls as the classical approaches due to their reliance on static training data, which is demonstrated in Table I. On comparing the performance of BERT to the classical ML models, BERT outperforms them marginally with a  \( \sim\ 3.5\) percent gain in accuracy.

\begin{table}[t]
\centering
\caption{Performance Metrics of our pipelines for each of the NLI models}
\begin{tabular}{|c|c|c|>{\centering\arraybackslash}p{13mm}|>{\centering\arraybackslash}p{13mm}|>{\centering\arraybackslash}p{13mm}|>{\centering\arraybackslash}p{13mm}|>{\centering\arraybackslash}p{13mm}|>{\centering\arraybackslash}p{13mm}|>{\centering\arraybackslash}p{13mm}|}
\cline{1-5}
\textit{\textbf{Pipeline}}& \textit{\textbf{Metrics}}& \textit{\textbf{FactCC}} & \multicolumn{2}{|c|}{\textit{\textbf{SummaC}}}\\
\cline{4-5}
  &  &  \textit{\textbf{}} & \textit{\textbf{Zero-Shot}}  & \textit{\textbf{Conv}}  \\ \cline{1-5}
  
 & Precision & 0.55 & 0.79 & 0.63  \\
\cline{2-5}
  \textit{\textbf{Question-Answer}} & Recall & 0.56 & 0.75 & 0.60  \\
\cline{2-5}
 \textit{\textbf{Pipeline}}& F1 & 0.55 & 0.77 & 0.62  \\
\cline{2-5}
 & Accuracy & 54.80\% & 77.60\%  & 62.50\% \\
 \hline

  & Precision & 0.54 & 0.77 & 0.69  \\
\cline{2-5}
  \textit{\textbf{SLM Pipeline}} & Recall & 0.47 & 0.74 & 0.57  \\
\cline{2-5}
 (Mistral-7B)& F1 & 0.50 & 0.75 & 0.63  \\
\cline{2-5}
 & Accuracy & 53.20\% & 75.80\%  & 65.80\% \\
\hline

  & Precision & 0.53 & 0.77 & 0.69  \\
\cline{2-5}
  \textit{\textbf{SLM Pipeline}} & Recall & 0.53 & 0.60 & 0.55  \\
\cline{2-5}
 (Phi-3)& F1 & 0.53 & 0.67 & 0.62  \\
\cline{2-5}
 & Accuracy & 53.28\% & 70.90\%  & 65.40\% \\
\hline

  & Precision & 0.78 & 0.90 & 0.96  \\
\cline{2-5}
  \textit{\textbf{Article Pipeline}} & Recall & 0.79 & 0.77 & 0.70  \\
\cline{2-5}
 & F1 & 0.78 & 0.83 & 0.81  \\
\cline{2-5}
 & Accuracy & 78.02\% & \textbf{84.30\%}  &83.60\% \\
 
\hline
\end{tabular}
\end{table}

\subsection{\textbf{How effective is our Proposed Solution for the given task?}}
Notably, our proposed solutions VERITAS-NLI, significantly outperforms prior approaches. From the results presented in Table III, The highest accuracy achieved from our proposed pipelines is \textit{84.3\%}, by the Article-Pipeline using SummaC-ZS as the NLI-based consistency checking model. This is a substantial \textit{33.3\%} gain over the best performing classical machine learning model (MultinomialNB) and a \textit{31.0\%} gain over the transformer-based BERT model. 
Out of the 12 pipeline configurations we tested (4 scraping approaches and 3 headline inconsistency detection NLI models), 10 configurations demonstrated superior accuracy compared to all baseline models.

The reliability of the solution proposed stems from our novel web-scraping techniques for dynamic information retrieval for the verification of the claim. With growing interpolation in writing techniques for articles our proposed solution captures, the discrete idea of the document and claim by making use of sentence level granularity. This also avoids having to retrain the model on newer data creating a robust solution for verification of claims.

Our solution provides explainability by providing the externally retrieved knowledge used to arrive at a decision as well as the links to Top-K articles fetched as displayed in Figure 5. This alleviates the black box created by other models and provides a way to interpret the models decision making process.

Table II elucidates the working of our proposed solution, with a detailed view of its performance for a given headline covering all pipeline approaches.

\begin{tcolorbox}[colback=gray!10, colframe=gray!50, boxrule=0.5mm, coltitle=black, fonttitle=\bfseries, sharp corners]
\textbf{Summary }: The Article-Pipeline with SummaC-ZS achieved the highest accuracy of \textit{84.3\%},surpassing the best classical ML model by \textit{33.3\%} and BERT by \textit{31\%}. Out of the 12 evaluated pipeline configurations - 10 of them outperformed all baselines in terms of accuracy.
\end{tcolorbox}

\begin{figure*}
    \centering
    \begin{tikzpicture}
        \node[draw=black, line width=0.1mm, rectangle, inner sep=10pt] (image) at (0,0) {
            \includegraphics[width=1\linewidth]{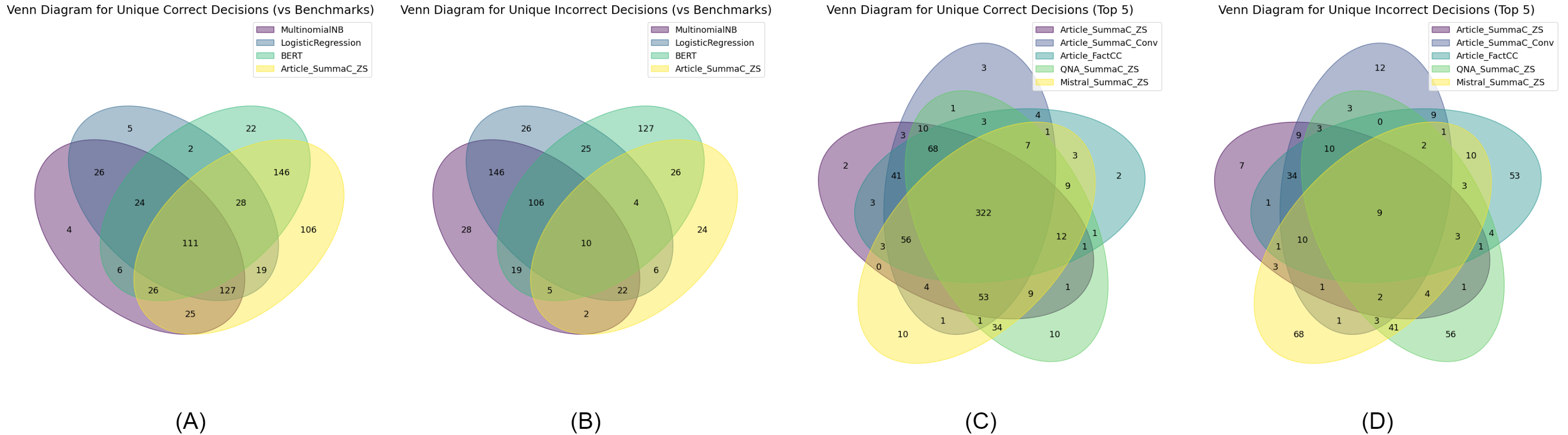}
        };
    \end{tikzpicture}
    \caption{Common and Unique Correct-Incorrect Decisions Between Pipelines}
    \label{fig:VennDiagram}
\end{figure*}

\subsection{\textbf{Which NLI-based Model performs best for Headline Inconsistency Detection?}}
For headline inconsistency detection, the SummaC model consistently outperforms FactCC, particularly due to its application of NLI-models on sentence-level pairs of a premise and hypothesis. SummaC is designed to capture fine-grained inconsistencies by analyzing individual sentence-level relations, while FactCC focuses on document-level fact-checking. This enhanced granularity allows SummaC to better detect subtle inconsistencies between headlines and their corresponding scraped content.

Moreover, SummaC's two variants—SummaC-ZS (zero-shot) and SummaC-Conv (Convolution)—outperform FactCC across all evaluation metrics. This superior performance can be attributed to SummaC's unique approach of leveraging a Natural Language Inference (NLI) pair-matrix. The NLI matrix enables SummaC to systematically compare sentence pairs and identify contradictions or entailments with greater precision, leading to more accurate headline inconsistency detection.

This is consistent with the results obtained in Table III, the average accuracy for FactCC, SummaC-ZS and SummaC-Conv based pipelines is \textit{59.83\%}, \textit{77.15\%} and  \textit{69.33\%} respectively. The SummaC pipelines outperform the FactCC pipelines by \textit{17.32\%} and \textit{9.50\%} each. In the case of SummaC-ZS, one drawback is its sensitivity to extreme values due mean calculation. However, since news headlines are predominantly one sentence, this potential drawback is no longer a hindrance to its performance for our use-case.

\begin{tcolorbox}[colback=gray!10, colframe=gray!50, boxrule=0.5mm, coltitle=black, fonttitle=\bfseries, sharp corners]
\textbf{Summary }: SummaC-ZS and SummaC-conv outperform FactCC  by \textit{17.32\%} and \textit{9.50\%} respectively in this task, due to their use of NLI models to analyze sentence-level premise and hypothesis pairs.
\end{tcolorbox}

\subsection{\textbf{Which Small Language Models (SLMs) perform best, and how effective are the generated questions? }}
In our study we make use of two small language models for the question generation step, the central idea was for the generated questions to help scrape more specific information needed to verify the claim.

The \texttt{Mistral-7b} pipelines have an average accuracy of \textit{64.93\%} and \texttt{Phi-3} pipelines have an accuracy of \textit{63.37\%}. This indicates that both these language models exhibit similar proficiency in the question generation task, despite Phi-3 having slightly over half the parameters compared to Mistral. However, the small language model pipelines only attain a maximum accuracy of \textit{75.80\%}, which is decisively beaten by the \textit{84.3\%} accuracy reached by the Article pipeline.

FLEEK~\cite{bib9} highlights that the quality of evidence retrieval is dependent on the initial set of questions generated by the language models as a potential limitation in their study,this is the probable factor for the degradation in performance compared to the naive article approach.

\begin{tcolorbox}[colback=gray!10, colframe=gray!50, boxrule=0.5mm, coltitle=black, fonttitle=\bfseries, sharp corners]
\textbf{Summary }: The \texttt{Mistral-7b} and \texttt{Phi-3} pipelines obtain average accuracies of \textit{64.93\%} and \textit{63.37\%} respectively, demonstrating similar performance despite Phi-3 having fewer parameters.However, these pipelines fall short compared to the \textit{84.3\%} accuracy of the Article pipeline.
\end{tcolorbox}
\subsection{\textbf{What are the performance characteristics and relationships between the Top-performing Classical models, BERT-based models and our Proposed Solution?}}
This research question aims to gain insights into the performance characteristics of the various legacy models, our proposed solution and to assess how independent or unrelated they are.

We take 2 different viewpoints to assess the degree of independence and utilize a Venn diagram to evaluate the number of distinct correct and incorrect decisions across these models and pipelines:
\begin{itemize}
    \item \textbf{Legacy Models vs. Our Best-Performing Pipeline} \\
    According to the results presented in Table I \& III, We pick the best performing classical ML models (Multinomial Naive Bayes, Logistic Regression and BERT-base-uncased) and compare them with our best-performing pipeline - Article Pipeline with SummaC Zero-Shot.\\
    From the venn diagrams in Figure 6A and 6B, we notice that the number of distinct correct decisions made by our solution far exceeds the corresponding numbers for the legacy models. Furthermore, our solution also has the lowest number of distinct incorrect decisions. Another insight lies in the significant overlap in incorrect decisions (Figure 6B) made by the legacy models, further highlighting the inability of models trained on static training data to generalize for the latest news headlines.
    \item \textbf{Comparison Between Our Best Performers}\\
    Figures 6C and 6D present a detailed comparison of correct and incorrect classifications produced by our best-performing pipelines. The Venn diagram emphasizes the distinctly correctly classified data points for the question-based pipelines, QNA\_SummaC\_ZS and Mistral\_SummaC\_ZS. Notably, both of these pipelines uniquely classifies 10 data points, with a total of 54 correct classifications that are not shared with other pipelines. In contrast, the article-based pipelines show far fewer unique correct classifications.
    This suggests that question-based pipelines are more effective at retrieving contextually valuable content in certain cases, particularly when addressing more nuanced questions about the claim. However, while QNA\_SummaC\_ZS and Mistral\_SummaC\_ZS exhibit the highest number of distinct correct classifications, they also account for the most distinct incorrect classifications. This indicates that while these pipelines excel in a specific category of summaries pertaining to 7.84\% of the evaluation dataset, they lack generalizability compared to the article-based pipelines, which perform more consistently across broader contexts. This further highlights that the quality of the content-retrieval is directly tied to the quality of the Question generated by the SLM  as highlighted in the above sections. 
    We also note that the Article\_SummaC\_ZS has significantly fewer distinct incorrect decisions than its counterpart Article\_FactCC which further cements its superiority for the task, due to its sentence-level granularity.
\end{itemize}
\begin{figure}
    \centering
    \begin{tikzpicture}
        \node[draw=black, line width=0.1mm, rectangle, inner sep=2pt] (image) at (0,0) {
            \includegraphics[width=1\linewidth]{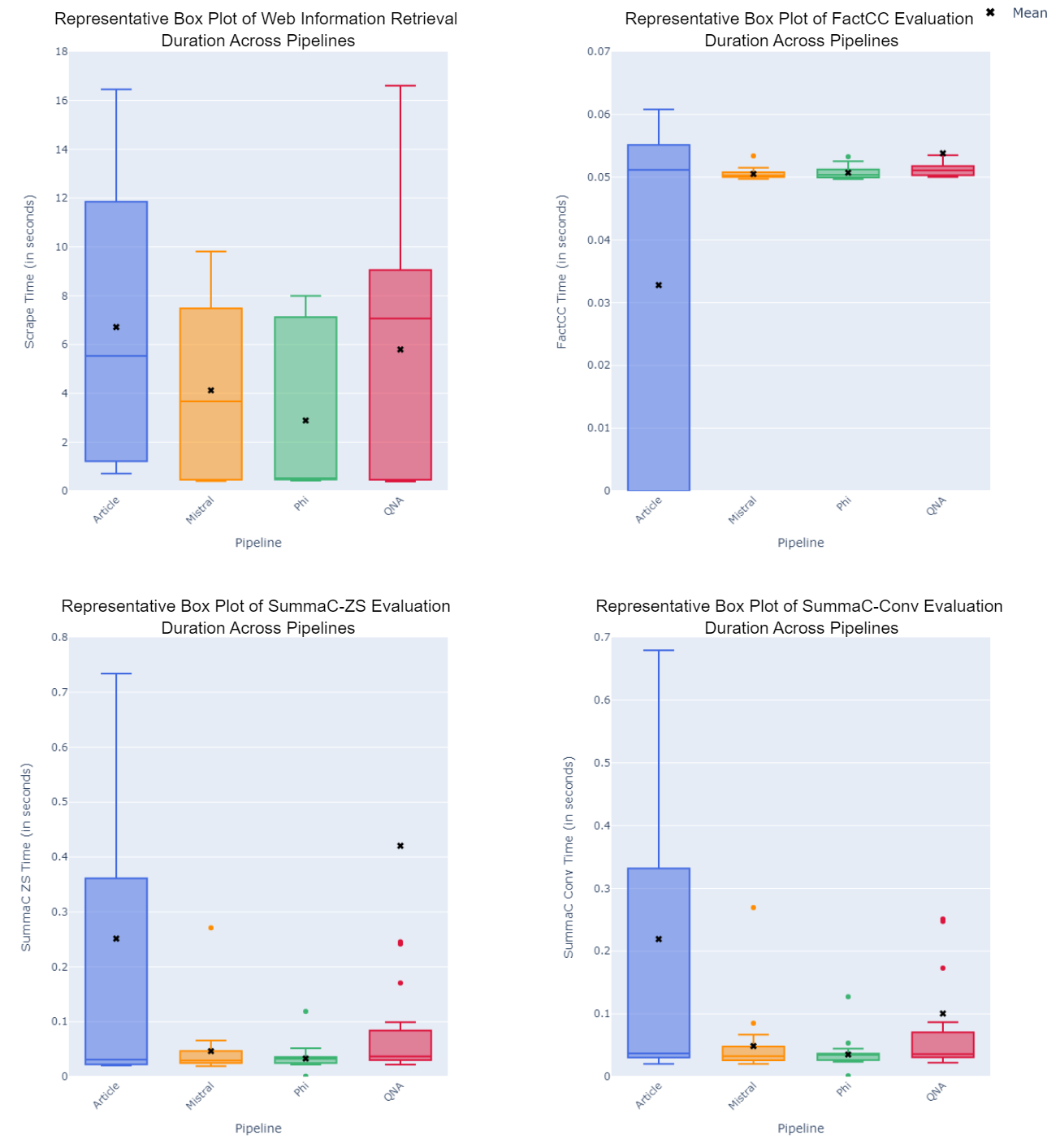}
        };
    \end{tikzpicture}
    \caption{Boxplot's illustrating execution times for various steps of our Proposed Pipeline.}
\end{figure}
\begin{tcolorbox}[colback=gray!10, colframe=gray!50, boxrule=0.5mm, coltitle=black, fonttitle=\bfseries, sharp corners] \textbf{Summary}: The best-performing classical models fall short of the unique correct and incorrect decisions metric compared to the Article Pipeline with SummaC Zero-Shot. The SummaC based Article pipelines make far fewer distinct incorrect decisions compared to their FactCC counterparts, highlighting their efficacy in detecting inconsistencies.\end{tcolorbox}

\subsection{\textbf{How efficient are our Proposed Pipelines? }}
To assess the performance of our solution, we conducted a representative analysis by calculating the average processing time. This was achieved by stratified sampling to form a slice of our evaluation dataset, which provided a balanced subset of true and false headlines. To ensure reproducibility and maintain consistency, all the experiments were performed on Google Colab with an NVIDIA Tesla T4 GPU (15.4GB GDDR6 memory) and Intel(R) Xeon(R) CPU @ 2.00GHz with 12.7GB RAM.\\ We divided our pipeline into two stages, the web-scraping stage and the NLI for inconsistency detection stage and evaluated them separately. As depicted in Figure 7, the time taken for each sub-task is visualised as a box-plot. Furthermore, Table IV contains  detailed information for each pipeline, well as the total duration inference for each.

The crucial task of scraping content to retrieve information from the web, is the most time-intensive, still averaging only \textit{2.9} seconds in the SLM-Phi3  pipeline, \textit{4.1} seconds in the SLM-Mistral pipeline, \textit{5.7} seconds for the Question-Answer pipeline and \textit{6.7} seconds for the Article pipeline. 

The next stage involves the use of NLI models to evaluate the support for the input headline (hypothesis) in the retrieved information (premise). We choose to examine this from two different perspectives :
\begin{itemize}
    \item \textbf{NLI-Model Specific Time Metrics}\\
    The average inference time for the NLI-models across our 4 pipelines is \textit{0.047s}, \textit{0.188s}, \textit{0.101s} for FactCC, SummaC-ZS and SummaC-Conv respectively.\\
    We observe that the time taken by the SummaC-ZS and SummaC-Conv models is greater than FactCC, this is due to the increased granularity of the SummaC Models as it segments the inputs into sentences that are individually compared. The increased time is a valuable trade off for a sizeable improvement in the accuracy of the model across the board as seen in Table III.
    \item \textbf{Pipeline Specific Time Metrics}\\
    Similarly the average inference times are \textit{0.168s}, \textit{0.049s}, \textit{0.040s} and \textit{0.192s} for the Article, SLM Mistral-7B, SLM Phi-3 and Question-Answer pipelines respectively. The average time taken is greatest for the Question-Answer and Article pipelines due to the larger volume of text retrieved by these 2 approaches, resulting in higher inference times.
\end{itemize}

\begin{tcolorbox}[colback=gray!10, colframe=gray!50, boxrule=0.5mm, coltitle=black, fonttitle=\bfseries, sharp corners] \textbf{Summary}: Our performance evaluation procedure for the 4 pipelines involves dividing the process into web scraping and NLI stages. Web scraping averaged 2.9-6.7 seconds, while NLI inference times ranged from 0.047s to 0.192s across models and pipelines. Larger text volumes in the Question-Answer and Article pipelines resulted in higher inference times overall. \end{tcolorbox}
\begin{table*}[htbp]
\centering
\caption{Mean Values of Time Metrics and Summation of Scrape, FactCC, SummaC ZS, and SummaC Conv Times for Each Pipeline}
\begin{tabular}{>{\centering\arraybackslash}p{0.15\linewidth} 
                >{\centering\arraybackslash}p{0.08\linewidth} 
                >{\centering\arraybackslash}p{0.08\linewidth} 
                >{\centering\arraybackslash}p{0.08\linewidth} 
                >{\centering\arraybackslash}p{0.08\linewidth} 
                >{\centering\arraybackslash}p{0.08\linewidth} 
                >{\centering\arraybackslash}p{0.08\linewidth} 
                >{\centering\arraybackslash}p{0.08\linewidth}}
\toprule
\textsf{\textbf{Pipeline}} & \textsf{\textbf{Scrape Time (s)}} & \textsf{\textbf{FactCC Time (s)}} & \textsf{\textbf{SummaC ZS Time (s)}} & \textsf{\textbf{SummaC Conv Time (s)}} & \textsf{\textbf{Sum of Scrape + FactCC Time (s)}} & \textsf{\textbf{Sum of Scrape + SummaC ZS Time (s)}} & \textsf{\textbf{Sum of Scrape + SummaC Conv Time (s)}} \\ 
\midrule
\textsf{\textit{\textbf{Article}}} & 6.7175 & 0.0328 & 0.2514 & 0.2193 & 6.7503 & 6.9689 & 6.9368 \\ 
\textsf{\textit{\textbf{SLM Mistral-7B}}} & 4.1240 & 0.0505 & 0.0464 & 0.0486 & 4.1745 & 4.1704 & 4.1726 \\ 
\textsf{\textit{\textbf{SLM Phi-3}}} & 2.8883 & 0.0507 & 0.0330 & 0.0352 & 2.9390 & 2.9213 & 2.9235 \\ 
\textsf{\textit{\textbf{Question-Answer}}} & 5.7986 & 0.0538 & 0.4205 & 0.1005 & 5.8524 & 6.2191 & 5.8991 \\ 
\bottomrule
\end{tabular}
\label{tab:time_metrics}
\end{table*}
\section{Conclusion}
In this study, we explore the efficacy of various Machine Learning and Transformer Models for the task of fake news detection, our findings highlight the inadequate performance characteristics of these classical models due to their reliance on static training data. Our proposed solution, VERITAS-NLI overcomes this limitation by combining advanced web-scraping techniques for the retrieval of external knowledge, required to verify a claim, and state-of-the-art Natural Language Inference Models to find support for the headline in the retrieved text. The results exhibit a significant improvement in effectiveness of our proposed solution in comparison to benchmark models, with gains of over \textit{30\%} in terms of accuracy.

These enhancements are crucial for advancing the field and ensuring that automated systems can keep pace with the continually evolving strategies employed in the dissemination of fake news, ultimately safeguarding the integrity of public discourse.
\section{Data Availability}
All datasets and code files used in our research study can be found in our Github Repository ~\cite{github}.
\bibliographystyle{ieeetr} 
\bibliography{ref}

\end{document}